\documentclass[letterpaper]{article} 
\usepackage[draft]{aaai25}
\usepackage{times}  
\usepackage{helvet}  
\usepackage{courier}  
\usepackage[hyphens]{url}  
\usepackage{graphicx} 
\usepackage{natbib}  
\usepackage{caption} 
\usepackage{algorithm}
\usepackage{algorithmic}
\usepackage{amsmath}
\usepackage{amssymb}

\usepackage{graphicx}
\usepackage{subcaption}
\usepackage{xcolor}

\usepackage{newfloat}
\usepackage{listings}
\DeclareCaptionStyle{ruled}{labelfont=normalfont,labelsep=colon,strut=off} 
\floatstyle{ruled}
\newfloat{listing}{tb}{lst}{}
\floatname{listing}{Listing}
\title{Adaptive Heavy-Tailed Stochastic Gradient Descent}

\author {
    Bodu Gong\textsuperscript{\rm 1}, 
    Gustavo Enrique Batista\textsuperscript{\rm 2}, 
    Pierre Lafaye de Micheaux\textsuperscript{\rm 1}
}
\affiliations {
    \textsuperscript{\rm 1}School of Mathematics and Statistics, University of New South Wales, Sydney, Australia \\
    \textsuperscript{\rm 2}School of Computer Science and Engineering, University of New South Wales, Sydney, Australia \\
    {bodu.gong, lafaye, gbatista}@unsw.edu.au
}

\usepackage{bibentry}

\begin{document}

\maketitle


\begin{abstract}
In the era of large-scale neural network models, optimization algorithms often struggle with generalization due to an overreliance on training loss. One key insight widely accepted in the machine learning community is the idea that wide basins (regions around a local minimum where the loss increases gradually) promote better generalization by offering greater stability to small changes in input data or model parameters. In contrast, sharp minima are typically more sensitive and less stable. Motivated by two key empirical observations - the inherent heavy-tailed distribution of gradient noise in stochastic gradient descent and the Edge of Stability phenomenon during neural network training, in which curvature grows before settling at a plateau, we introduce Adaptive Heavy Tailed Stochastic Gradient Descent (AHTSGD). The algorithm injects heavier-tailed noise into the optimizer during the early stages of training to enhance exploration and gradually transitions to lighter-tailed noise as sharpness stabilizes. By dynamically adapting to the sharpness of the loss landscape throughout training, AHTSGD promotes accelerated convergence to wide basins. AHTSGD is the first algorithm to adjust the nature of injected noise into an optimizer based on the Edge of Stability phenomenon. AHTSGD consistently outperforms SGD and other noise-based methods on benchmarks like MNIST and CIFAR-10, with marked gains on noisy datasets such as SVHN. It ultimately accelerates early training from poor initializations and improves generalization across clean and noisy settings, remaining robust to learning rate choices.
\end{abstract}

\section{Introduction}
Optimization is a cornerstone of modern scientific and engineering practice, which drives advances in fields such as computational chemistry, engineering design, and financial modeling \cite{10.1007/978-3-7908-2604-3_16}. In the age of artificial intelligence, optimization plays an especially critical role in training neural networks, where it is used to minimize a prescribed loss function and identify the most informative model parameters. The objective is to find an optimal set of parameters \( \mathbf{w} \in \mathbb{R}^N \) that minimizes an empirical loss function \( f(\mathbf{w}) \).   

\begin{figure*}[htb] 
    \centering

    \begin{subfigure}[t]{0.49\linewidth}
        \centering
        \includegraphics[width=\linewidth]{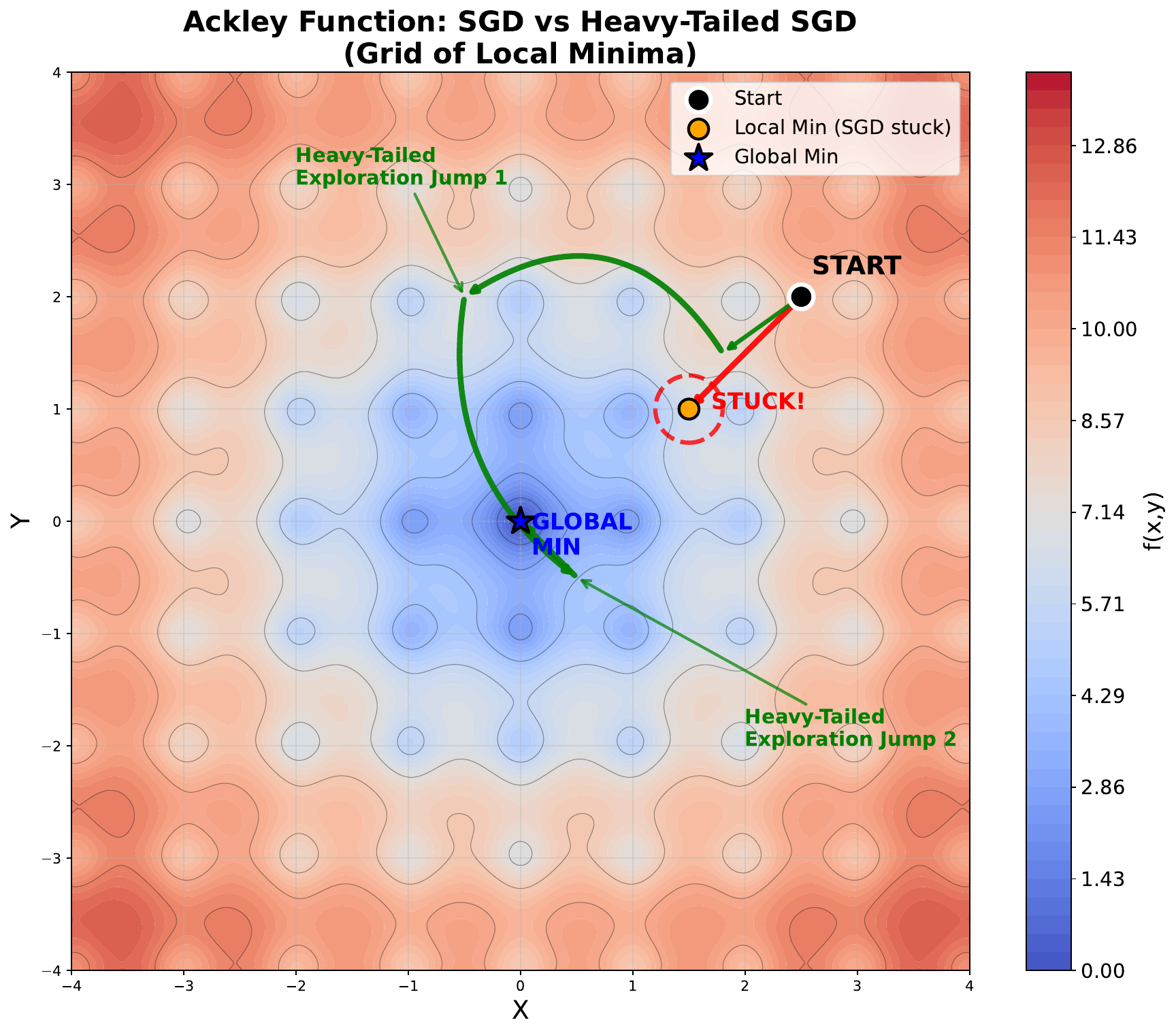}
        \caption{Ackley Function}
        \label{fig:ackley}
    \end{subfigure}
    \hfill
    \begin{subfigure}[t]{0.49\linewidth}
        \centering
        \includegraphics[width=\linewidth]{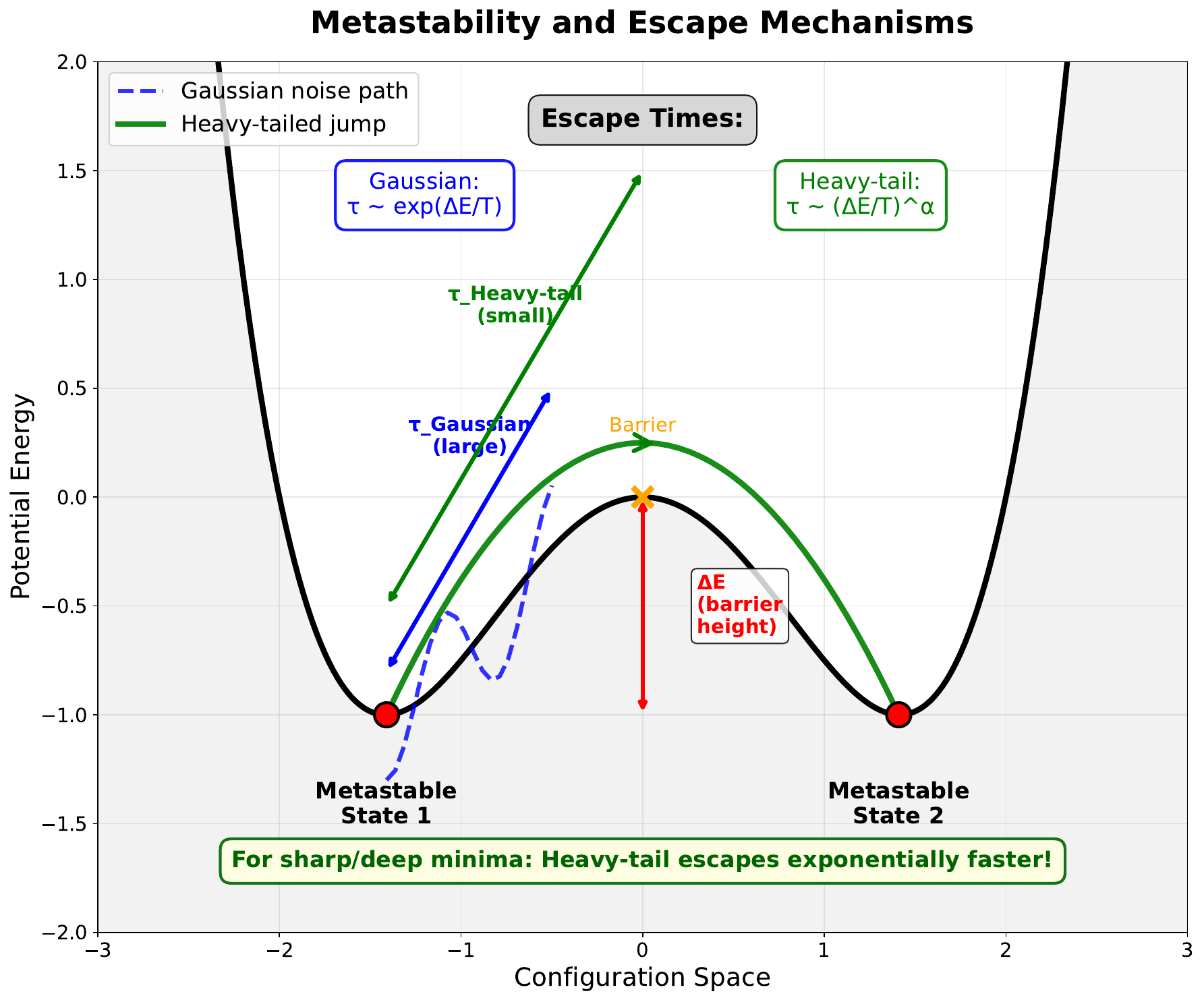}
        \caption{Metastability Analysis}
        \label{fig:metastability}
    \end{subfigure}



    \caption{(a) Ackley Function heavy tailed noise encourage exploration in early stages of training and thus experience better results in the early stages. (b) Metastability analysis of SGD with Lévy $\alpha$-stable noise compared to Gaussian noise, showing that the former contains higher probability of escaping sharp local minima.}
    \label{fig:all_landscape_trajectories}
\end{figure*} 

Among the myriad of classes of optimization techniques, gradient-based methods, which iteratively update parameters using local derivative information, form the backbone of training procedures in deep learning~\cite{lecun2015deep}. We have gradient descent: 

\begin{equation}
\theta_{t+1} = \theta_t - \eta \nabla_\theta \mathcal{L}(\theta_t)
\end{equation}

where $\theta_t$ denotes the model parameters at iteration $t$, $\eta$ is the learning rate, and $\nabla_\theta \mathcal{L}(\theta_t)$ is the gradient of the loss function $\mathcal{L}$ with respect to the parameters $\theta$ at iteration $t$. One of the most influential methods in this class is Stochastic Gradient Descent (SGD)~\cite{kiefer-wolfowitz}, which approximates the full-batch gradient using random subsets of data, significantly enhancing scalability. 

\begin{equation}
\theta_{t+1} = \theta_t - \eta \nabla_\theta \mathcal{L}(\theta_t; x_t, y_t)
\end{equation}

where $\nabla_\theta \mathcal{L}(\theta_t; x_t, y_t)$ represents the gradient of the loss function with respect to $\theta_t$, evaluated at the training example $(x_t, y_t)$. The widespread adoption of SGD has catalyzed breakthroughs in tasks like image classification and natural language processing, and continues to be the de facto optimizer in many deep learning pipelines~\cite{ruder-gradient-descent}. However, despite its simplicity and effectiveness, SGD suffers from several key limitations, most notably, its tendency to get trapped in sharp, narrow local minima or saddle points, which may impair generalization to unseen data~\cite{dauphin2014identifying, jastrzebski2017three}.

A promising strategy to mitigate this issue involves injecting noise into the optimization process. This not only aids in escaping suboptimal basins but also encourages convergence to wider, flatter minima which are regions of low curvature that have been empirically linked to improved generalization performance~\cite{hochreiter1997flatminima, keskar2016largebatch, zhou2019noise}. Conventional approaches such as Stochastic Gradient Langevin Dynamics (SGLD)~\cite{welling2011sgld} introduce Gaussian noise, justified by the Central Limit Theorem. 

\begin{equation}
\theta_{t+1} = \theta_t - \eta \nabla_\theta \mathcal{L}(\theta_t; x_t, y_t) + \sqrt{2 \eta} \epsilon_t
\end{equation}

where $\epsilon_t \sim \mathcal{N}(0, I)$ represents Gaussian noise sampled from a standard normal distribution. However, recent empirical studies reveal that the noise structure in SGD does not conform to Gaussianity. Instead, the noise exhibits heavy-tailed behaviour, better described by Lévy \( \alpha \)-stable distributions, where \( \alpha \) denotes the tail index~\cite{simsekli2019tail}. Unlike Gaussian noise, Lévy noise permits occasional large jumps, offering a more powerful mechanism to explore complex loss landscapes and avoid getting trapped in narrow valleys. Thus stochastic gradient descent with Lévy \( \alpha \)-stable distributed noise.

\begin{equation}
\theta_{t+1} = \theta_t - \eta \nabla_\theta \mathcal{L}(\theta_t; x_t, y_t) + \eta^{1/\alpha} \cdot \xi_t
\end{equation}

where $\xi_t \sim \mathcal{S}_\alpha(0, 1, 0)$ is a symmetric Lévy $\alpha$-stable random variable, and the scaling term $\eta^{1/\alpha}$ ensures that the noise is appropriately tempered relative to the learning rate. 


We propose \textbf{Adaptive Heavy-Tailed Stochastic Gradient Descent (AHTSGD)}, a geometry-aware optimizer that dynamically adjusts the noise tail index \( \alpha \) based on the evolving sharpness of the loss landscape. Motivated by the observation that the leading Hessian eigenvalue initially rises before plateauing near the edge of stability, AHTSGD uses this signal to modulate exploration: it applies heavier-tailed noise (\( \alpha < 2 \)) when sharpness increases, enabling escape from narrow minima, and gradually shifts toward Gaussian noise as curvature stabilizes. Unlike methods such as EntropySGD or SAM, AHTSGD requires no inner-loop optimization or extra hyperparameters, offering a lightweight alternative for sharpness-aware training.

Our key contributions are threefold:
\begin{enumerate}
    \item \textbf{Adaptive noise framework:} We introduce a principled Lévy \( \alpha \)-stable noise algorithm that dynamically modulates its tail index based on the leading Hessian eigenvalue, aligning the noise distribution with evolving loss sharpness.
    \item \textbf{Empirical validation:} Across synthetic landscapes and standard deep-learning benchmarks, AHTSGD consistently escapes narrow minima more quickly and converges more reliably than existing optimisers.
    \item \textbf{Robust to poor initialization and learning-rate agnosticism:} We show that AHTSGD converges rapidly even when weights begin extremely close to zero, scenarios that challenge most methods, and that its performance remains stable across a wide range of learning rates, eliminating the need for expensive rate tuning.
\end{enumerate} 

By bridging heavy-tailed stochasticity with sharpness awareness, AHTSGD opens a new avenue for understanding and improving the optimization of deep neural networks.

\section{Method}

We begin by demonstrating the power of heavy-tailed noise using the classic Ackley function—a multimodal benchmark with global minimum at \( \mathbf{x}^* = (0, \ldots, 0) \), where \( f(\mathbf{x}^*) = 0 \). It combines a radially symmetric decay term \( \exp\left(-20 \sqrt{\frac{1}{N} \sum x_i^2} \right) \) with an oscillatory component \( \cos(2\pi x_i) \), producing sharp local minima spaced approximately one unit apart. Owing to its rotational symmetry \cite{pinnau2017consensus}, the Ackley function is widely used to evaluate optimizer robustness to initialization, given its rugged yet structured landscape \cite{noack2017hybrid, back1996evolutionary}.

To ensure fair comparison, we apply uniformly scaled noise sampled from a symmetric Lévy distribution with stability index \( \alpha = 1.5 \). Figure~\ref{fig:ackley} illustrates the optimization trajectories. Standard SGD frequently becomes trapped in sharp local minima or fails to traverse critical regions. In contrast, Lévy \(\alpha\)-stable SGD enables more robust exploration, escaping suboptimal basins and converging to flatter or global minima that generalize better. Notably, in Figure~\ref{fig:ackley}, Lévy-driven SGD escapes symmetric traps where SGD remains stuck, highlighting its superior exploratory capability over longer horizons.

\section{Sharpness Dynamics and the Edge of Stability}

\begin{figure*}[htb]
    \centering
    \includegraphics[width=\linewidth]{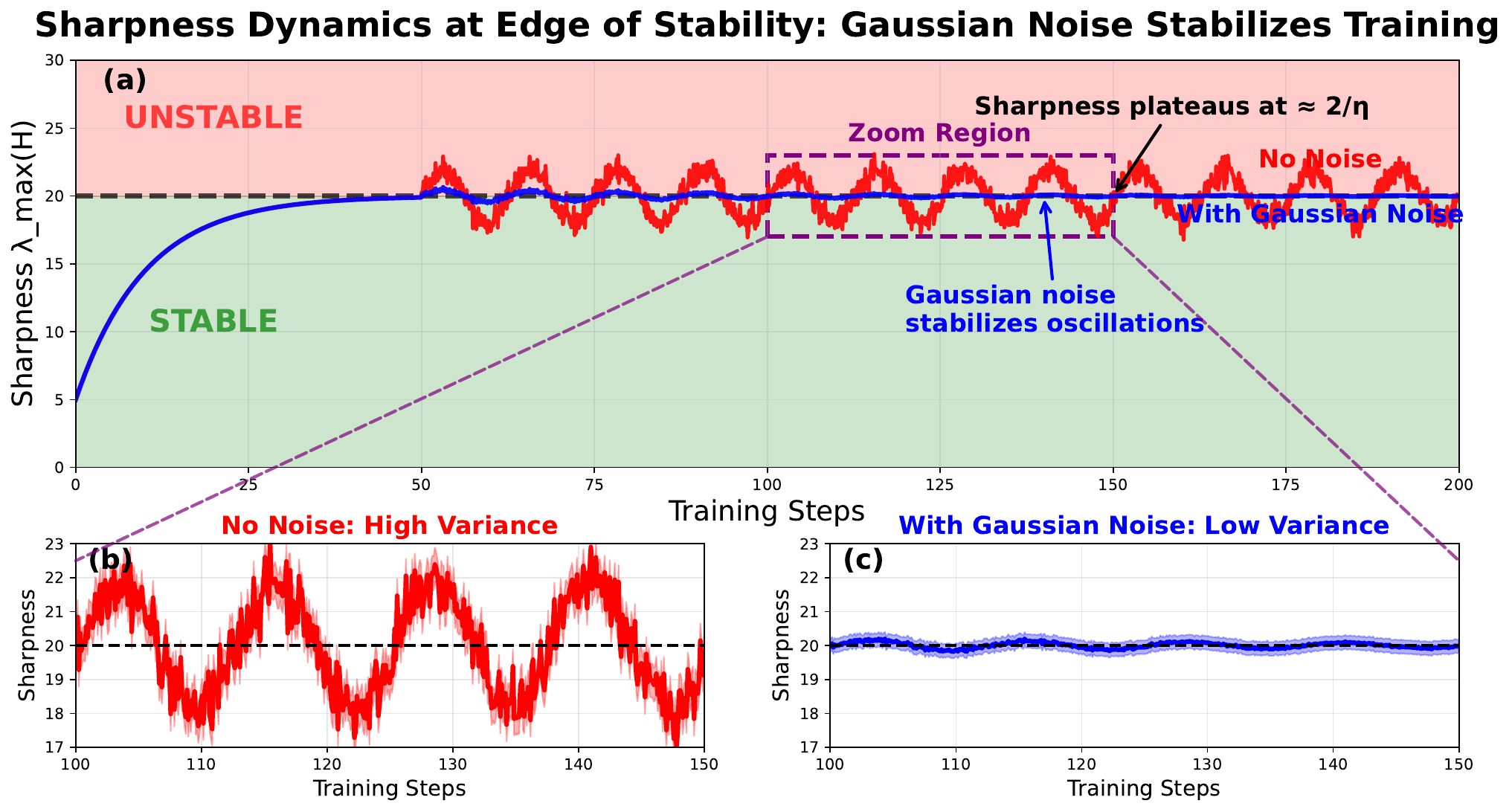}
    \caption{(a) Evolution of sharpness \(\lambda_{\max}(H)\) during training, showing that sharpness rises and plateaus near the critical threshold \(\frac{2}{\eta}\). Without noise (red), oscillations are large and persist in the unstable region. Gaussian noise (blue) reduces these oscillations and stabilizes training dynamics. (b) Close-up view of sharpness under no noise: high variance and sustained oscillations. (c) Close-up view of sharpness under Gaussian noise: significantly reduced variance and damped oscillations.}
    \label{fig:plateau}
\end{figure*}

While Lévy $\alpha$-stable noise encourages exploration, it can also introduce instability if not properly controlled. To make exploration both effective and stable, we propose an adaptive mechanism for $\alpha$ based on the evolving geometry of the loss landscape. Specifically, we track sharpness, defined as the largest eigenvalue of the Hessian, ($\lambda_{\max}$), throughout training where $\lambda_{\max}$ fluctuates during early training but eventually stabilizes near a plateau as shown in Figure \ref{fig:plateau}. This plateau often aligns with the edge of stability, approximately \( \lambda_{\max} \approx \frac{2}{\eta} \) where $\eta$ is the learning rate and learning dynamics of SGD become sensitive to noise \cite{andreyev2025edge}.

We interpret this sharpness plateau as a real-time indicator of the local curvature: a high $\lambda_{\max}$ suggests a sharp and narrow region of the loss surface, while lower values indicate flatter and wide-basin regions. Standard SGD injects isotropic Gaussian noise and cannot adapt to this evolving landscape \cite{feng2021inverse}. Fixed heavy-tailed noise can help escape sharp minima, but without decay or adaptation, it can destabilize convergence in flatter regions. 

Our solution is to adapt the tail index $\alpha_t$ of Lévy $\alpha$-stable noise based on the evolving sharpness $\lambda_{\max}(t)$. In sharp regions (near the plateau), we reduce $\alpha_t$ to inject heavier-tailed noise, enabling the optimizer to make large, rare jumps that facilitate escape. In flatter regions, where such volatility is unnecessary, we increase $\alpha_t \to 2$ to recover the stability of Gaussian noise.

This intuition is supported by metastability theory for Lévy-driven SDEs \cite{simsekli2019tail}, where the expected escape time from a sharp local minimum with energy barrier $H$ scales as:
\begin{equation}
\mathbb{E}[T_{\text{escape}}] \sim \exp\left(\frac{H}{\sigma_t^{\alpha_t}}\right).
\end{equation}

where $H$ is the energy barrier height and $\sigma_t$ is the decaying noise scale as identified in Figure \ref{fig:metastability}. Compared to Gaussian noise ($\alpha = 2$), heavy-tailed noise with $\alpha_t < 2$ yields:

\begin{equation} \frac{\mathbb{E}[T_{\text{escape}}^{\text{SGD}}]}{\mathbb{E}[T_{\text{escape}}^{\text{AHTSGD}}]} = \exp\left(\frac{H}{\sigma^2} - \frac{H}{\sigma^{\alpha_t}} \cdot \Gamma(1 + 2/\alpha_t)^{\alpha_t/2}\right) \end{equation}

which grows rapidly in sharp regions where $\alpha_t \to 1$. 


\subsection{Adaptive Alpha} \label{sec:AlphaUpdate}
We introduce the crux of the our adaptive algorithm, updating the alpha depending on the window of sharpness values that allows it to decide whether to increase or decrease. We keep track of the exponentially smoothed sharpness values (denoted as EMA\_Sharpness), which capture the evolving sharpness associated with the Edge of Stability phenomenon. This allows us to capture the overall trend while avoiding sensitivity to short-term fluctuations. Additionally, we track the current value of the stability parameter $\alpha$, which is adapted dynamically as optimization progresses.

Across multiple simulations, we observed that the trajectory of the smoothed sharpness, denoted $\bar{s}_t$, closely follows a sigmoidal progression over the course of training. To exploit this empirical regularity, we apply a sigmoid transformation of the form:

\begin{equation}
z_t = \frac{1}{1 + \exp(-v(\bar{s}_t - c))},
\end{equation}

The parameter \( v \) controls the steepness of the sigmoid transition, while \( c \) determines the midpoint at which the transition occurs. This transformation produces a continuous value \( z_t \in [0, 1] \), which serves as a normalized indicator of local curvature. To integrate this curvature-aware signal into our algorithm, we linearly map \( z_t \) to the permissible range of the Lévy tail index \( \alpha \in [1, 2] \) as follows:
\begin{equation}
\alpha_t^{\text{raw}} = \alpha_{\min} + (\alpha_{\max} - \alpha_{\min}) \cdot z_t,
\end{equation}
where we fix \( \alpha_{\min} = 1 \) and \( \alpha_{\max} = 2 \). This linear mapping allows us to adaptively control the tail-heaviness of the Lévy \(\alpha\)-stable noise in response to the curvature of the loss landscape.

The output of this equation, denoted as $\alpha_t^{\text{raw}}$, is bounded within the interval $[1, 2]$. This ensures that as the sharpness increases during training in accordance with the Edge of Stability phenomenon, our $\alpha$ increases alongside it ensuring that in earlier stages of the optimization, we have a lower $\alpha$ promoting exploratory dynamics, while towards the end of the edge of stability phenomenon we yield lighter-tailed (more Gaussian-like) noise ($\alpha \approx 2$), promoting convergence stability. 

Finally, to avoid abrupt changes in our $\alpha$ parameter, we employ an incremental update mechanism: 
\begin{equation} 
\alpha \leftarrow \alpha + \lambda \cdot (\alpha_t^{\text{raw}} - \alpha) 
\end{equation}
where $\lambda$ is a smoothing coefficient, fixed at 0.1 based on empirical tuning. This update rule ultimately facilitates stable adaptation of $\alpha$, ensuring that gradual shifts align with the increase in sharpness in training dynamics. 

In particular, we demonstrate that when the sharpness measure, defined as the maximum eigenvalue of the Hessian, satisfies $\lambda_{\max} > \frac{2}{\eta}$, the error bound achieved by AHTSGD is provably tighter than that of standard SGD. As the learning rate $\eta$ increases, this inequality is more likely to be satisfied, aligning with the Edge of Stability regime. Under this regime, our method adaptively constricts the error bound by annealing the variance of the heavy-tailed noise, transitioning toward Gaussian noise. This results in improved convergence behavior in sharper regions of the loss landscape.

\begin{algorithm}[tb]
\caption{Adaptive Heavy Tailed Stochastic Gradient Descent (AHTSGD)}
\label{alg:AHTSGD}
\textbf{Input}: Initial parameters $\theta$, learning rate $\eta$, loss function $\ell(\theta)$, training data $(x_i, y_i)_{i=1}^n$\\
\hspace*{1.2em} Noise scale $\text{noise\_init}$, noise decay rate $\gamma$, logistic params $v, c$, exponential annealing decay $k$ \\
\hspace*{1.2em} $\alpha_{\min}=1$, $\alpha_{\max}=2$, device $\in$ \{CPU, GPU\}, adaptive flag, $\rho=0.05$\\
\hspace*{1.2em} (optional) momentum, weight decay\\
\textbf{Output}: Optimized parameters $\theta_K$
\begin{algorithmic}[1]
\STATE Initialize $\theta$, $\alpha \gets \alpha_{\min}$, $\text{EMA\_sharpness} \gets 0$
\FOR{$k = 1$ to $K$}
    \FOR{each mini-batch $\mathcal{D}_k$}
        \STATE $\ell_k \gets \ell(\theta; \mathcal{D}_k)$       \COMMENT{Compute mini-batch loss}
        \STATE $s_t \gets \log(1 + |\text{Tr}(\nabla^2 \ell_k)|)$    \COMMENT{Estimate sharpness using Hutchinson’s method}
        \STATE $\text{EMA\_sharpness} \gets (1 - \rho) \cdot \text{EMA\_sharpness} + \rho \cdot s_t$
        \IF{adaptive}
            \STATE $z \gets \frac{1}{1 + \exp(-v(\text{EMA\_sharpness} - c))}$
            \STATE $\alpha_t^{\text{raw}} \gets \alpha_{\min} + (\alpha_{\max} - \alpha_{\min}) \cdot z$
            \STATE $\alpha \gets \alpha + \lambda(\alpha_t^{\text{raw}} - \alpha)$
        \ELSE
            \STATE $\alpha \gets 2 - \exp(-k \cdot x)$
        \ENDIF
        \STATE $\sigma \gets \frac{\sqrt{\text{noise\_init}}}{(1 + t)^{\gamma / 2}}$
        \IF{device is GPU}
            \STATE Sample $L \sim \text{Chambers-Mallows-Stuck}(\alpha, \beta=0, \sigma, \mu=0)$
        \ELSE
            \STATE Sample $L \sim \text{LevyStable}(\alpha, \beta=0, \sigma, \mu=0)$
        \ENDIF
        \STATE Update $\theta \gets \theta - \eta \cdot \nabla \ell_k + \eta^{1/\alpha} \cdot L$
    \ENDFOR
    \STATE Optionally evaluate model performance
\ENDFOR
\STATE \textbf{return} $\theta$
\end{algorithmic}
\end{algorithm}

We formalize this intuition with the following upper bound on the expected suboptimality:
\begin{multline}
\mathbb{E}[f(\theta_t) - f(\theta^*)] \leq 
\frac{L\|\theta_0 - \theta^*\|^2}{2t} + \frac{\eta L \sigma^2}{2} \cdot \\
\begin{cases}
1 & \text{(SGD)} \\
\left(\dfrac{\lambda_{\max}}{2/\eta}\right)^{2 - \alpha_t} & \text{(AHTSGD)}
\end{cases}
\end{multline}

Here, $L$ is the smoothness constant, $\sigma^2$ denotes the variance of the injected noise, and $\alpha_t$ is the time-varying stability parameter in AHTSGD. The term $\left(\frac{\lambda_{\max}}{2/\eta}\right)^{2-\alpha_t}$ captures how the algorithm adaptively modulates noise based on the curvature, effectively tightening the error bound as the optimization progresses into sharper regions.





\subsection{Annealing Alpha} \label{sec:AlphaUpdate}

We now present a simplified variant of the Adaptive Alpha strategy. While the full Adaptive Alpha method constructs a moving window and maintains an exponential moving average (EMA) of sharpness values, this simplified approach avoids the need for historical tracking. 

Instead, we enforce a smooth exponential transition from Lévy $\alpha$-stable noise to Gaussian noise by annealing the stability parameter $\alpha_t$ over time. Specifically, we define:
\begin{equation}
\alpha_t = 2 - \exp(-k t)
\end{equation}
where \( k \) is a tunable hyperparameter that governs the rate of decay.  This formulation ensures that $\alpha_t$ evolves within the range $[1, 2]$, starting from $\alpha_0 = 1$ and asymptotically approaching $\alpha = 2$ as training progresses.

This schedule provides a principled and computationally efficient means of modulating the tail-heaviness of the noise based on training time, enabling strong exploratory behavior during early optimization phases and gradually favoring more stable, Gaussian-like updates. Empirically, we observe that typical training regimes, such as MNIST with batch size 64, consist of approximately 900 mini-batches per epoch, which allows sufficient time for this transition to occur in alignment with the curvature dynamics of the loss landscape.



We demonstrate that in this Annealing Alpha setting (where the noise scale decays exponentially with rate $\gamma$ and $C_\alpha$ is a constant dependent on the Lévy distribution) the AHTSGD algorithm achieves faster convergence, particularly when the tail index $\alpha_t < 2$ during the early iterations. Specifically, we establish the following bound on the expected squared distance to the optimum:
\begin{equation} \mathbb{E}[\|\theta_T - \theta^*\|^2] \leq \frac{\|\theta_0 - \theta^*\|^2}{T^{\gamma}} + \frac{C_\alpha}{T^{\min(\gamma, 1-1/\alpha(T))}} \sum_{t=1}^T \frac{\sigma_t^{\alpha_t}}{t^{\gamma}} 
\end{equation}

This result highlights that when $\alpha_t < 2$, the heavy-tailed noise can accelerate exploration and yield improved convergence in the early stages of training, especially in sharp or rugged regions of the loss landscape.   

Furthermore, a key result that we theorise, under the annealing schedule $\alpha_t = 2 - e^{-kt}$ with noise scale $\sigma_t = \frac{\sigma_0}{(1+t)^{\gamma/2}}$, AHTSGD satisfies:
\begin{equation}
\mathbb{E}[|\theta_T - \theta|^2] \leq \frac{|\theta_0 - \theta|^2}{T^{\gamma}} + \frac{C_\alpha}{T^{\min(\gamma, 1 - 1/\alpha(T))}} \sum_{t=1}^T \frac{\sigma_t^{\alpha_t}}{t^{\gamma}},
\end{equation}
where $C_\alpha = \mathbb{E}[|L_\alpha|^2]$.For standard SGD (i.e., $\alpha = 2$),
\begin{equation}
\mathbb{E}[|\theta_T^{\text{SGD}} - \theta|^2] \leq \frac{|\theta_0 - \theta|^2}{T^{\gamma}} + \frac{\sigma^2}{T^{\gamma}} \sum_{t=1}^T \frac{1}{t^{\gamma}}.
\end{equation}

The relative improvement due to AHTSGD in early iterations is given by:
\begin{equation}
\mathcal{R}_{\text{early}} = \frac{\text{SGD Error}}{\text{AHTSGD Error}} = T^{\gamma - \min(\gamma, 1 - 1/\alpha(T))}.
\end{equation}

Thus, when \(\alpha(T) < 2 - \gamma\), we obtain \(\mathcal{R}_{\text{early}} > 1\), confirming the presence of early-stage acceleration. This implies a faster rate of improvement during the initial phase of training compared to standard SGD with Gaussian noise.

Ultimately, this adaptive strategy not only offers enhanced convergence speed and stability but also provides a flexible framework to adjust exploration dynamically based on the intrinsic sharpness characteristics of the optimization trajectory. Our findings strongly suggest reconsidering traditional fixed schedules for learning rate and noise perturbations, instead advocating adaptive methods responsive to the evolving geometry of the loss landscape, especially at critical regions identified through empirical observations such as the \textit{Edge of Stability}.

\begin{figure*}[t]
    \centering
    \begin{subfigure}[t]{0.38\textwidth}
        \centering
        \includegraphics[width=\linewidth]{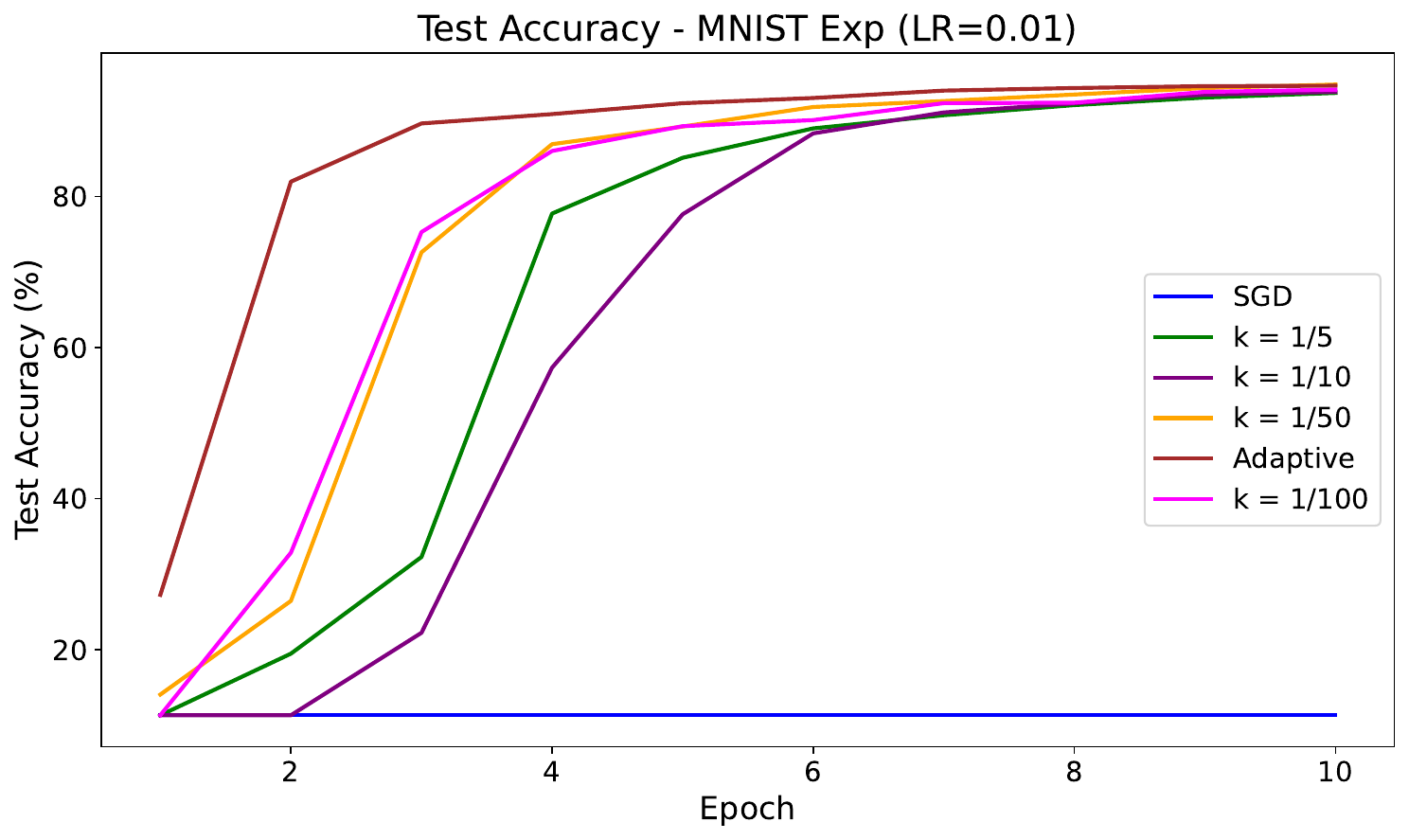}
        \caption{Zero Init, LR = 0.01}
        \label{MNIST_0}
    \end{subfigure}
    \begin{subfigure}[t]{0.38\textwidth}
        \centering
        \includegraphics[width=\linewidth]{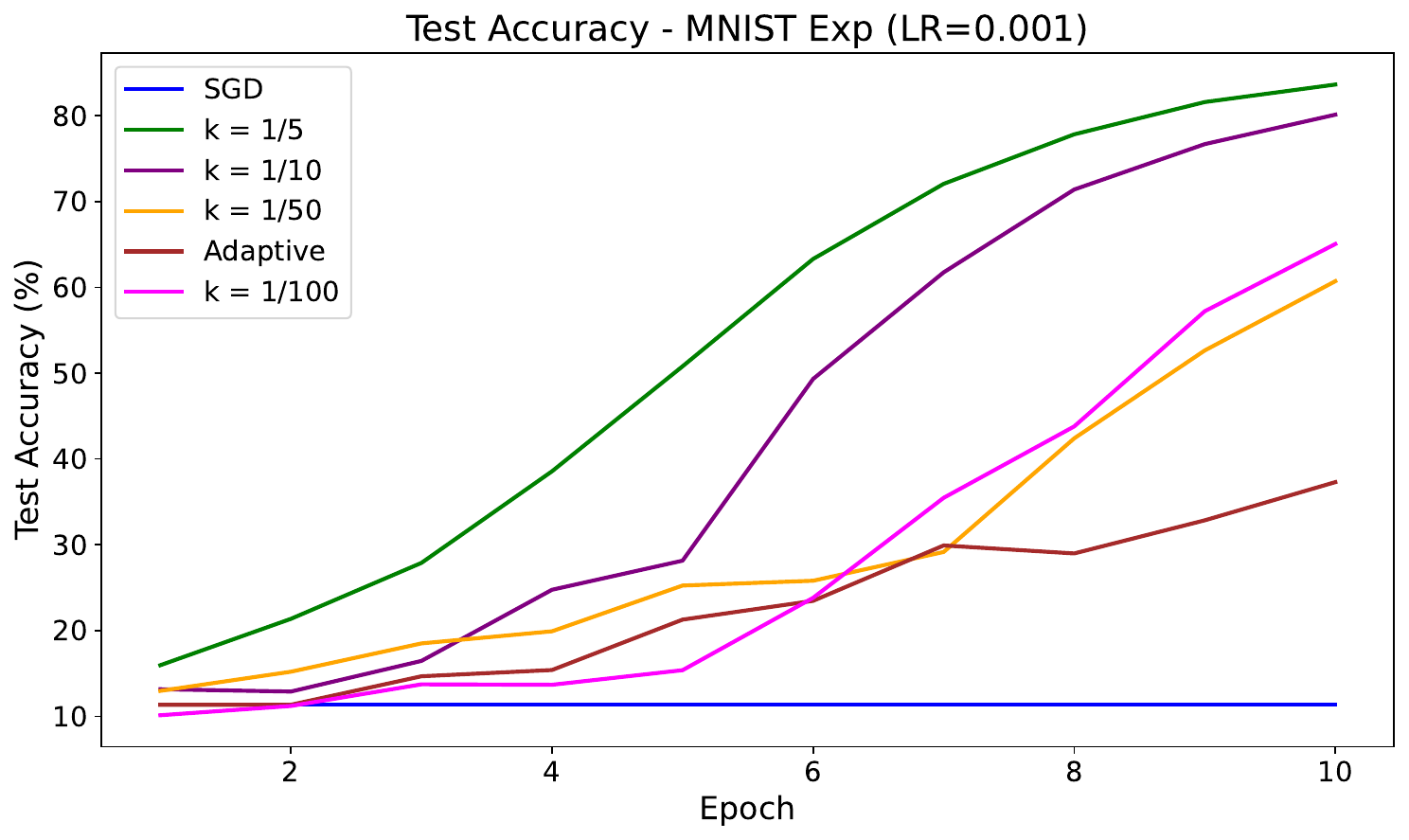}
        \caption{Zero Init, LR = 0.001}
        \label{MNIST_1}
    \end{subfigure}
    
    \vspace{1em}
    
    \begin{subfigure}[t]{0.38\textwidth}
        \centering
        \includegraphics[width=\linewidth]{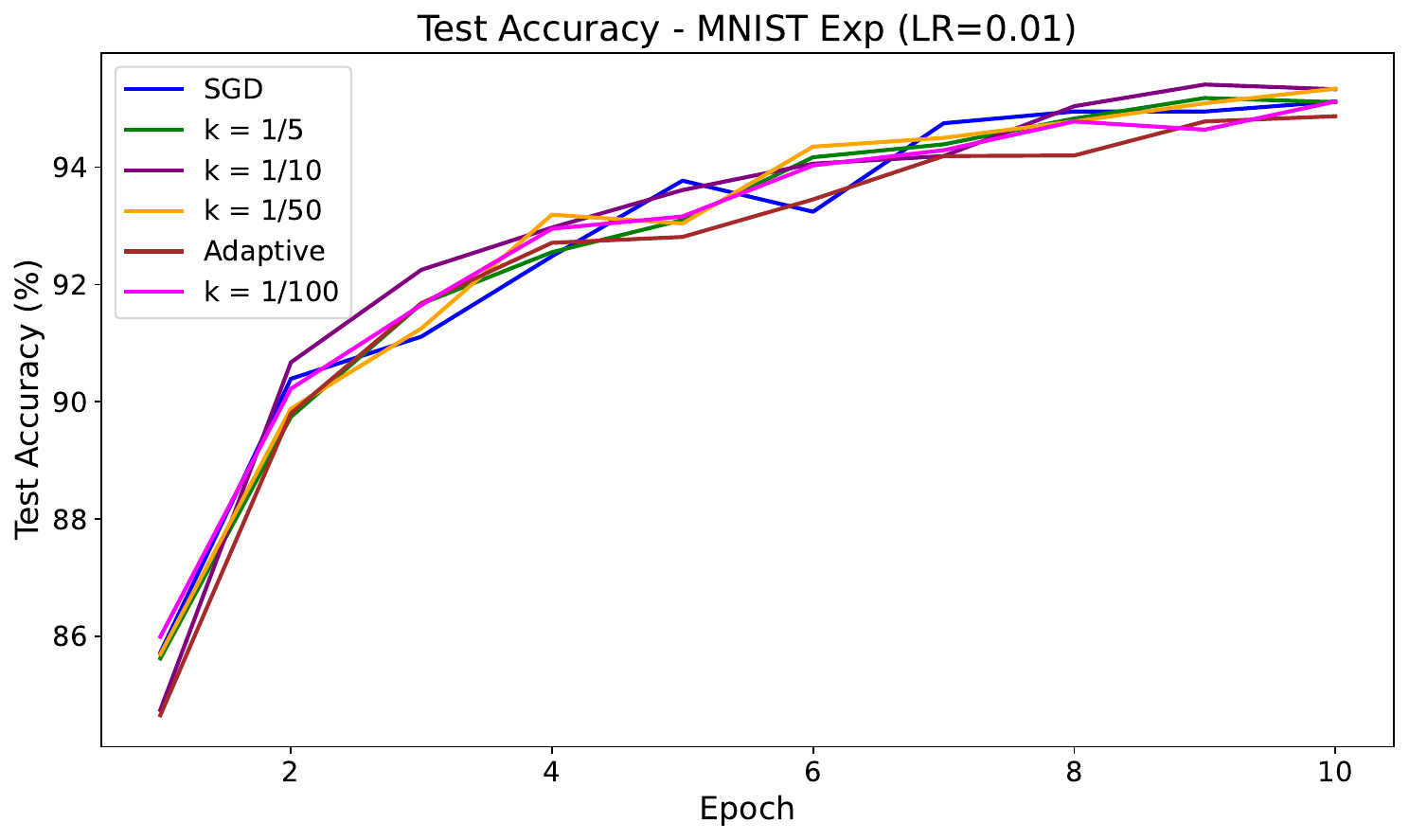}
        \caption{Kaiming Init, LR = 0.01}
        \label{MNIST_2}
    \end{subfigure}
    \begin{subfigure}[t]{0.38\textwidth}
        \centering
        \includegraphics[width=\linewidth]{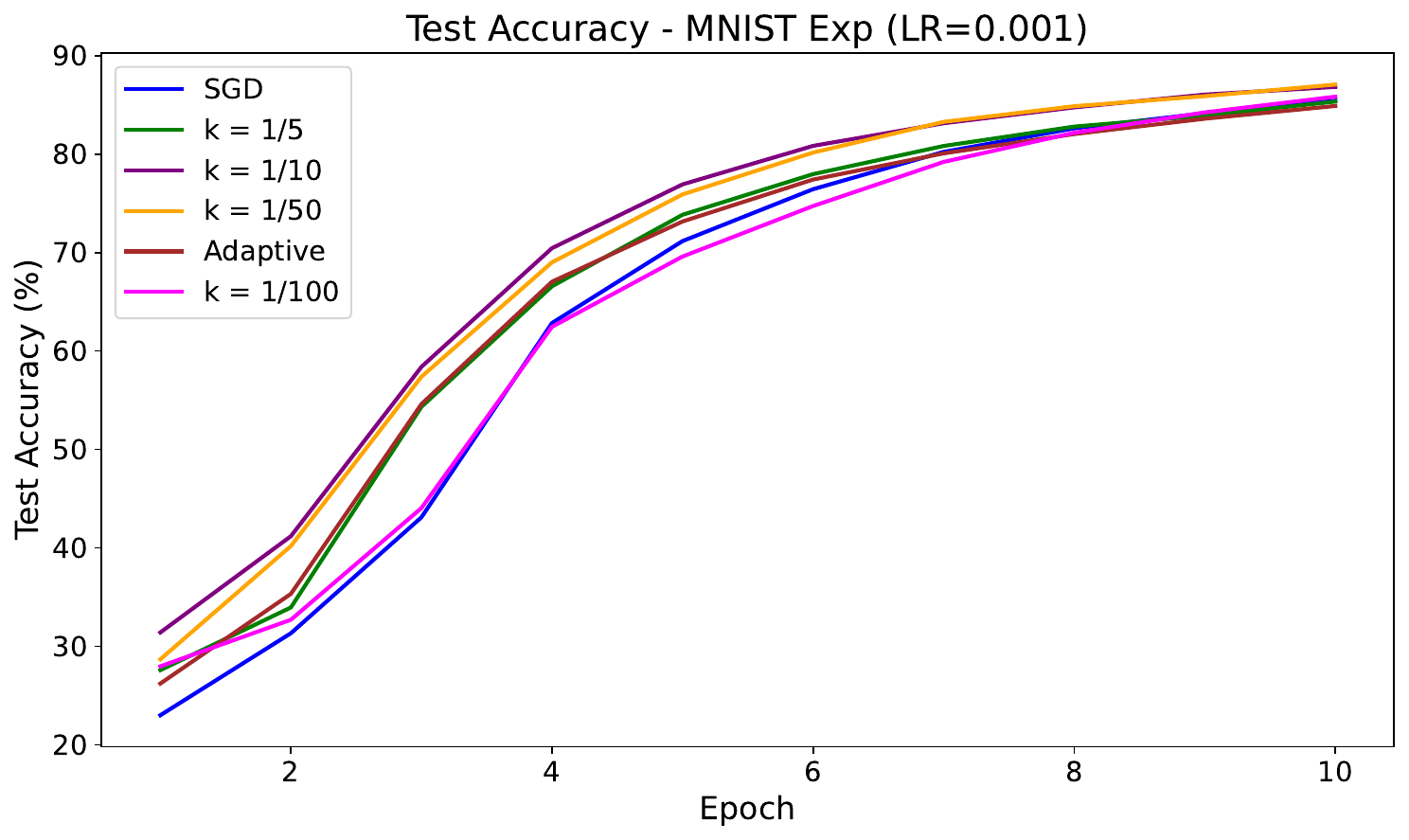}
        \caption{Kaiming Init, LR = 0.001}
        \label{MNIST_3}
    \end{subfigure}
    
    \caption{MNIST test accuracy using a 3-layer MLP under different weight initializations and learning rates.}
    \label{fig:MNISTInitialisationComparison}
\end{figure*}

\begin{figure*}[t]
    \centering
    \begin{subfigure}[t]{0.38\textwidth}
        \centering
        \includegraphics[width=\linewidth]{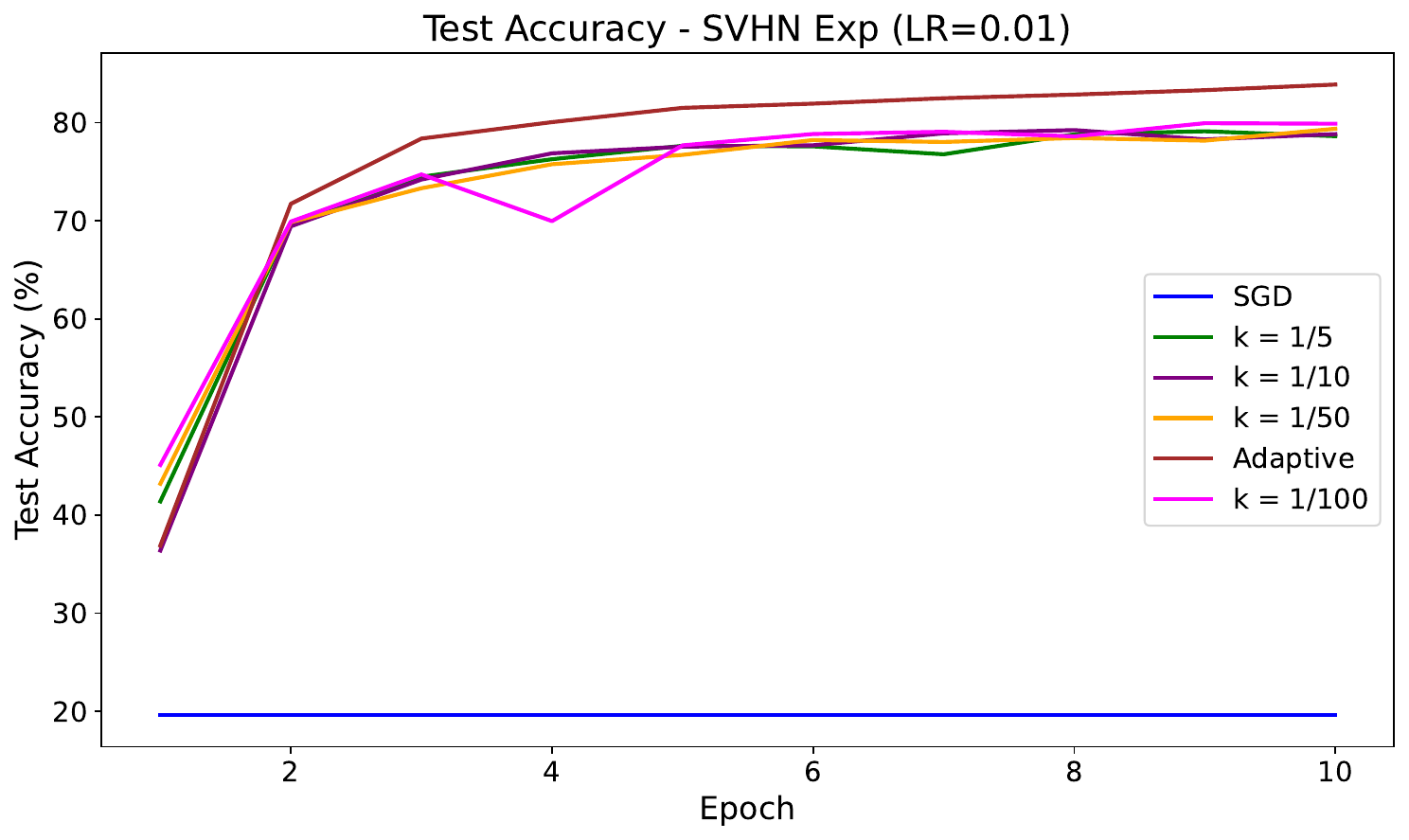}
        \caption{Zero Init, LR = 0.01}
        \label{SVHN_0}
    \end{subfigure}
    \begin{subfigure}[t]{0.38\textwidth}
        \centering
        \includegraphics[width=\linewidth]{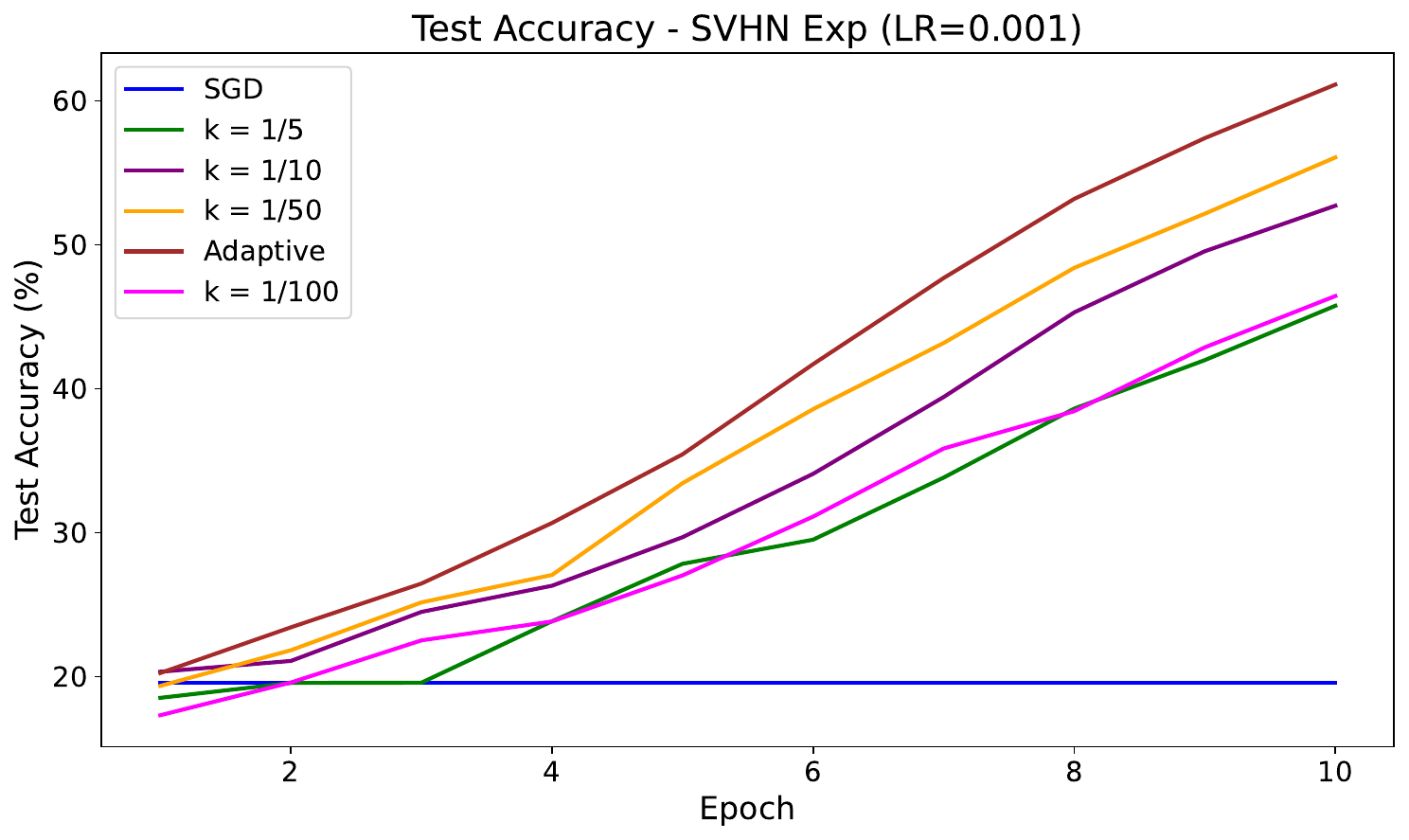}
        \caption{Zero Init, LR = 0.001}
        \label{SVHN_1}
    \end{subfigure}
    
    \vspace{1em}
    
    \begin{subfigure}[t]{0.38\textwidth}
        \centering
        \includegraphics[width=\linewidth]{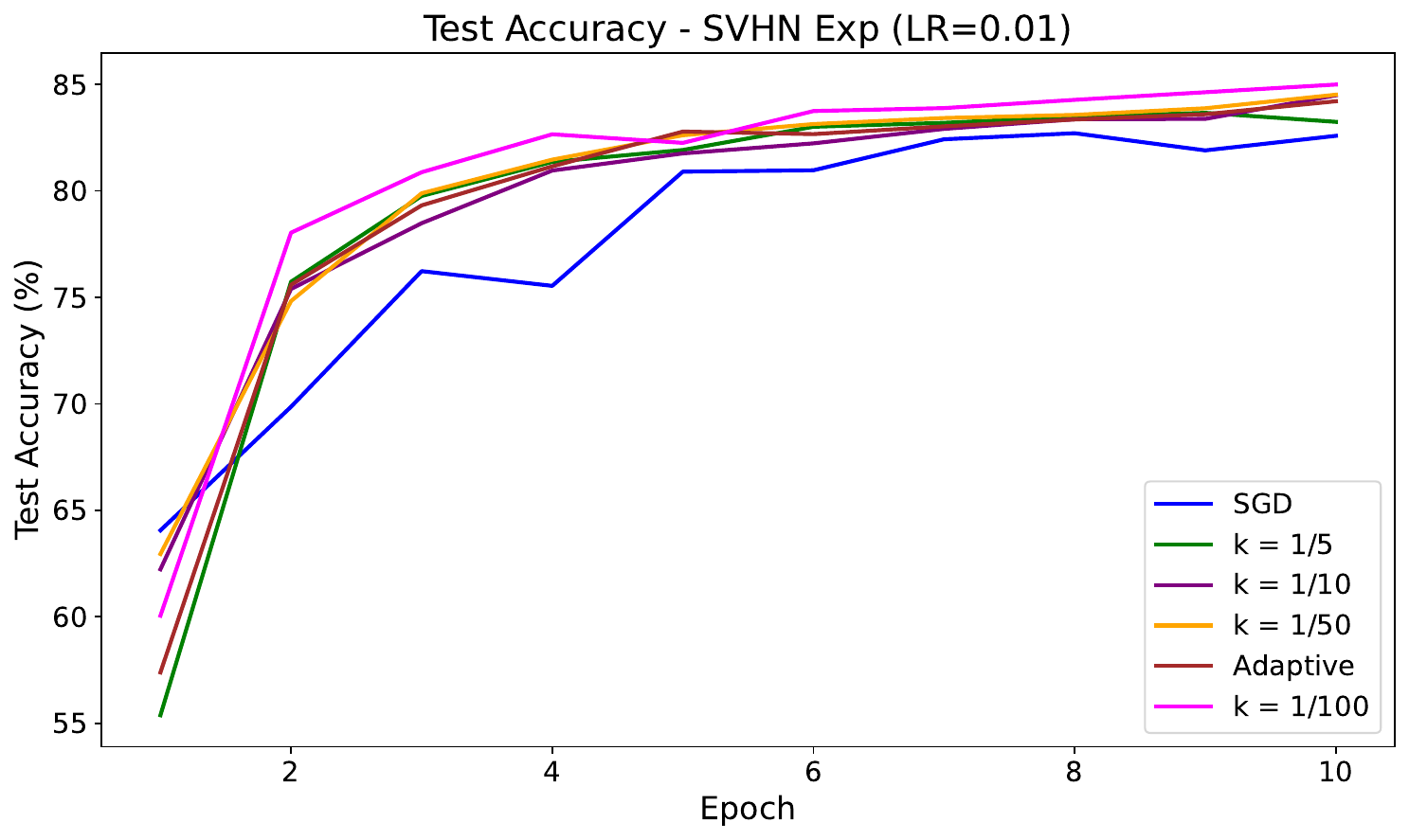}
        \caption{Kaiming Init, LR = 0.01}
        \label{SVHN_2}
    \end{subfigure}
    \begin{subfigure}[t]{0.38\textwidth}
        \centering
        \includegraphics[width=\linewidth]{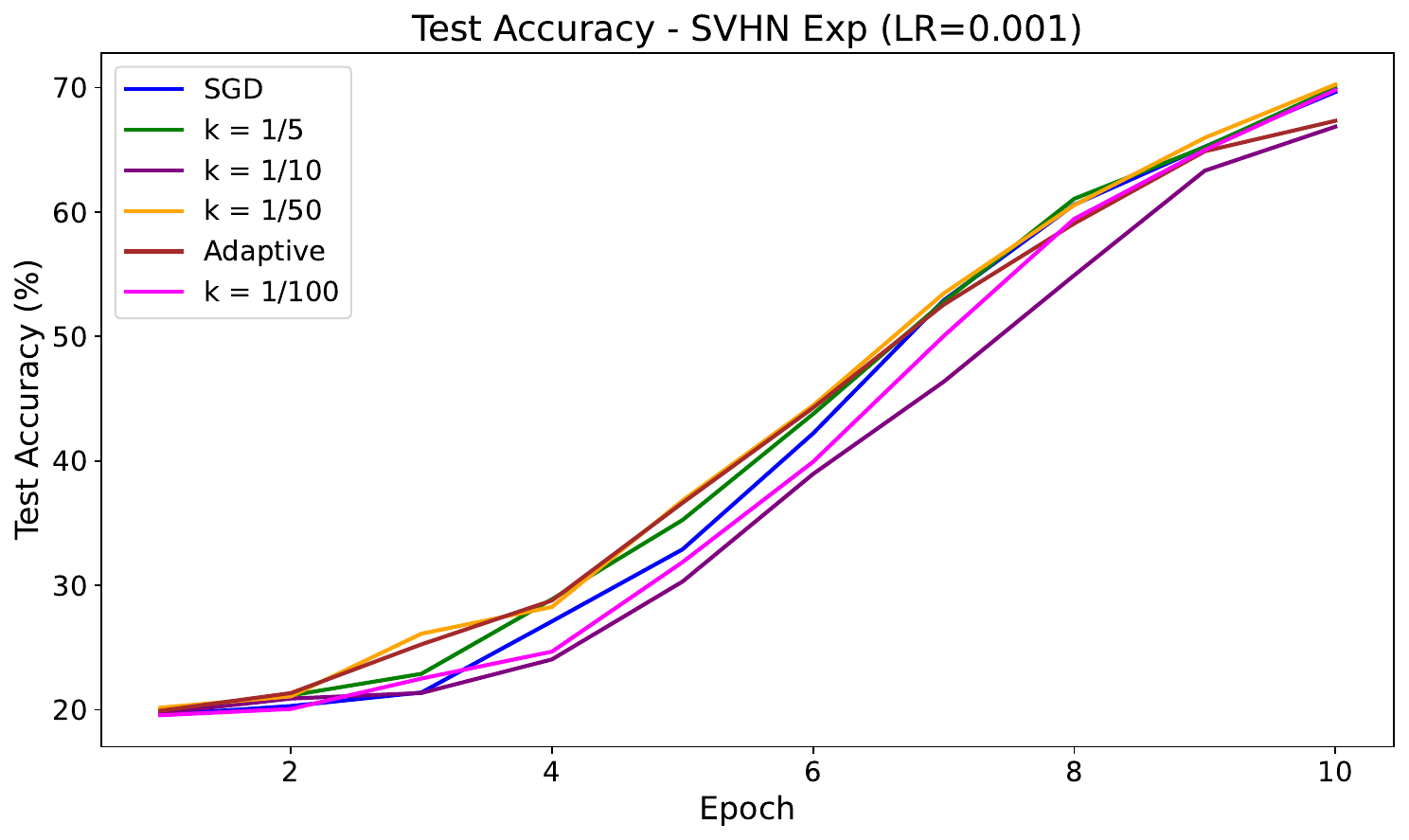}
        \caption{Kaiming Init, LR = 0.001}
        \label{SVHN_3}
    \end{subfigure}
    
    \caption{SVHN test accuracy using a CNN under different weight initializations and learning rates.}
    \label{fig:SVHNInitialisationComparison}
\end{figure*}

\subsection{Computational Efficiency: Sampling and Sharpness Estimation} \label{sec:CMSSample}






To scale AHTSGD efficiently to deep networks, we implement two key computational improvements. First, we employ the Chambers–Mallows–Stuck (CMS) method to sample from Lévy $\alpha$-stable distributions directly on GPU. Since these distributions lack closed-form densities for general $\alpha$, CMS provides an efficient and numerically stable procedure. All required random variables - uniform \( V \sim \mathcal{U}(-\frac{\pi}{2}, \frac{\pi}{2}) \) and exponential \( W \sim \mathrm{Exp}(1) \) are generated on GPU to support parallel, per-parameter noise injection at each iteration.

Second, to estimate the dominant eigenvalue of the Hessian, we use Hutchinson’s stochastic approximation with only 1–3 samples per step. This low-cost method effectively tracks sharpness trends throughout training and informs the adaptive update of the stability parameter \(\alpha_t\).

Together, these optimizations enable AHTSGD to remain computationally efficient while scaling to high-dimensional deep learning tasks.

\section{Experiment}

\subsection{Experimental Setup and Benchmarks} \label{sec:setup}
We conduct extensive experiments comparing our annealing-based and adaptive alpha heavy-tailed optimization methods against standard SGD across a range of canonical benchmarks. These include image classification tasks on MNIST and SVHN using multi-layer perceptrons, as well as ResNet-50 on CIFAR-10 dataset.

For all experiments, we fix learning rates $\eta \in \{10^{-2}, 10^{-3}\}$ (except ResNet-50, where we only use $\eta = 10^{-2}$ due to its depth), and explore two values for the noise initialization parameter: $\{0.001, 0.005\}$, while holding the decay rate constant at $0.55$ following \cite{neelakantan2017adding}. The non-adaptive variants are trained using exponential schedules for $\alpha$ with $k \in \left\{\frac{1}{5}, \frac{1}{10}, \frac{1}{50}, \frac{1}{100}, \frac{1}{1000} \right\}$. AHTSGD, denoted as \textit{adaptive} in the plots, adjusts $\alpha$ based on landscape sharpness.

To focus purely on the effects of the optimizer and noise schedule, we avoid additional techniques such as learning rate scheduling, dropout, batch normalization, or weight decay, except for ResNet-50, where momentum and weight decay are included for convergence stability. All models are trained with standard data augmentation (random cropping, horizontal flipping, 4-pixel padding) and averaged over 15 random seeds.

\subsection{Three-layer MLP on MNIST} \label{sec:MLPResults}

We evaluate the initialization invariance and convergence behavior of AHTSGD on a 3-layer MLP with ReLU activations and no dropout. We consider both zero initialization and Kaiming uniform initialization \cite{he2015delving}. Models are trained on the MNIST dataset with batch size 64 for training and 1000 for testing. Standard normalization is applied using mean = 0.1307 and std = 0.3081 \cite{Piciarelli2019CapsuleAnomaly, Thiruthummal2024InfoGeometry}.

\subsection{CNN on SVHN}

We next evaluate performance on the SVHN dataset using a simple CNN with two convolutional layers followed by ReLU and max pooling, and a final fully connected classification layer. The model is trained using the same optimizer configurations as MNIST, and comparisons are again made between zero and Kaiming initializations. No dropout or batch normalization is used. The training batch size is 64 and testing batch size is 1000. Input images are normalized using SVHN-specific means and standard deviations: mean = [0.4377, 0.4438, 0.4728], std = [0.1980, 0.2010, 0.1970] \cite{scellier2023energy}.

\subsection{CIFAR-10 with ResNet-50}

To evaluate performance on deeper architectures, we train ResNet-50 on CIFAR-10. Due to its scale (23.5 million parameters), we use a fixed learning rate of 0.01, weight decay of $5 \times 10^{-4}$, and momentum of 0.9, following guidelines from \cite{wu2022selecting}. We apply Kaiming uniform initialization and standard CIFAR-10 normalization (mean = [0.4914, 0.4822, 0.4465], std = [0.247, 0.243, 0.261]) \cite{ramakrishnan2019differentiable}. The training batch size is 128 and testing batch size is 1000. No learning rate scheduling is used.



We observe a striking acceleration in test accuracy on both MNIST and SVHN, particularly under challenging conditions. As shown in Figures~\ref{MNIST_0}, \ref{SVHN_0}, \ref{MNIST_1}, and \ref{SVHN_1}, when our 3 layered MLP (MNIST task) and 2 convolution layered CNN initialized with zeros, AHTSGD, both in its adaptive and annealing forms, successfully trains and achieves significantly higher performance, while standard SGD fails to converge. Furthermore, even with basic Kaiming Uniform initialization (without using more sophisticated schemes such as Xavier), AHTSGD still outperforms SGD, highlighting its robustness.

The benefits of AHTSGD become more pronounced as the difficulty of the task increases. For instance, in MNIST (Figures~\ref{MNIST_2} and \ref{MNIST_3}), the performance gap is less dramatic at lower learning rates (e.g., $0.0001$), but becomes clear at higher learning rates such as $0.001$. On the more challenging and noisier SVHN dataset, AHTSGD consistently outperforms SGD even under basic initialization, reinforcing its early-stage acceleration properties and initialization invariance.

\begin{figure}[h]
    \centering
    \includegraphics[width=\linewidth]{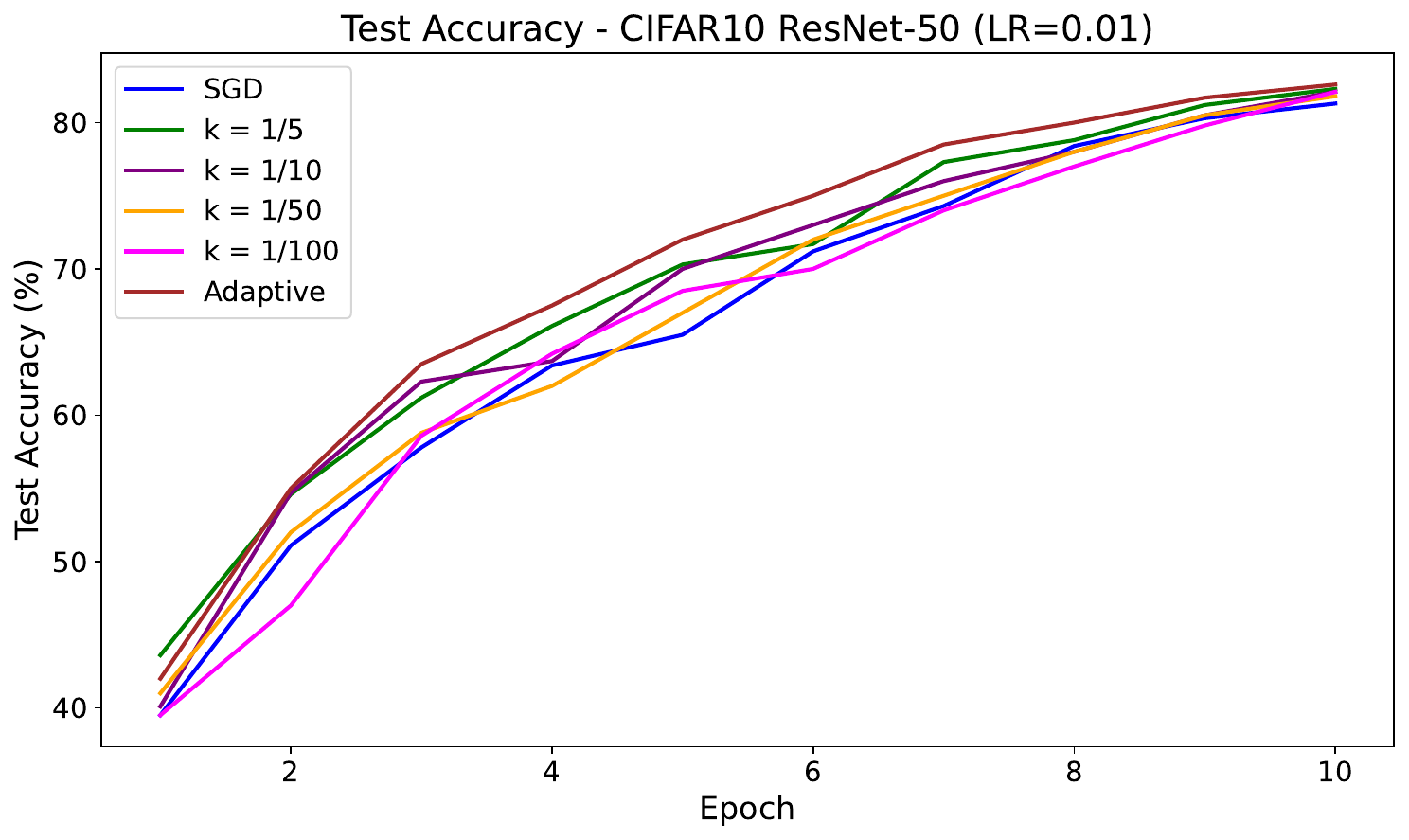}
    \caption{CIFAR-10 test accuracy using ResNet-50 (Kaiming Init, LR = 0.01).}
    \label{fig:KaiUnifCIFAR10LR0.01}
\end{figure}

This phenomenon persists in deeper architectures. For example, in Figure~\ref{fig:KaiUnifCIFAR10LR0.01}, we demonstrate that ResNet-50 trained on CIFAR-10 with Kaiming Uniform initialization also benefits significantly from AHTSGD.

Since the variance of injected noise in SGD naturally decays over mini batch iterations, its impact diminishes over time. Thus, our focus is placed on the first few epochs, where the dynamics of noise and sharpness most strongly influence convergence.

\section{Related Work}
This work focuses on sharpness and noise-aware first-order optimization methods. Sharpness-Aware Minimization (SAM) \cite{foret2020sharpness} minimizes the worst-case loss within an $\ell_2$-neighborhood of the parameters, encouraging convergence to flatter minima with better generalization. Adaptive SAM (ASAM) \cite{zhuang2022adaptive} extends this by scaling the neighborhood radius according to the geometry of each layer, improving performance in anisotropic landscapes.Furthermore, noise-based optimizers such as EntropySGD \cite{EntropySGD} introduce stochasticity to encourage exploration of flatter regions in the loss landscape, serving as a classic example of noise-aware optimization.


\section{Conclusion}

We introduced Adaptive Heavy-Tailed Stochastic Gradient Descent (AHTSGD), which dynamically adjusts the tail index $\alpha$ of Lévy $\alpha$-stable noise based on loss landscape sharpness. By employing heavier-tailed noise ($\alpha < 2$) during early training when sharpness increases, and transitioning to Gaussian-like noise ($\alpha \to 2$) as sharpness stabilizes, AHTSGD balances exploration and exploitation without additional hyperparameters or inner-loop optimization.

Our experiments demonstrate three key advantages: (1) \textbf{accelerated convergence}, with 5-20\% test accuracy improvements in early epochs across MNIST, SVHN, and CIFAR-10; (2) \textbf{initialization robustness}, successfully training even from zero initialization where SGD fails; and (3) \textbf{learning rate invariance}, maintaining stable performance across different learning rates. These benefits are particularly pronounced on noisy datasets like SVHN, where adaptive heavy-tailed noise effectively navigates complex loss landscapes.

AHTSGD represents a practical step toward geometry-aware optimization, automatically adapting the nature of noise injection to the Edge of Stability phenomenon without manual intervention.

\bibliography{aaai25}

\end{document}